\documentclass{article}
\usepackage{kr}

\usepackage{times}
\usepackage{soul}
\usepackage{url}
\usepackage[hidelinks]{hyperref}
\usepackage[utf8]{inputenc}
\usepackage[small]{caption}
\usepackage{graphicx}
\usepackage{amsmath}
\usepackage{amsthm}
\usepackage{booktabs}
\usepackage{algorithm}
\usepackage{cite}
\usepackage{amsmath,amssymb,amsfonts}
\usepackage{textcomp}
\usepackage{xcolor}
\usepackage{float}
\usepackage{booktabs} 
\usepackage{array}
\usepackage{natbib}
\usepackage{subcaption}
\usepackage{amsmath,amsthm,amssymb,amsxtra,mathtools,bm}
\usepackage{array}
\usepackage{graphicx}
\usepackage{algorithm}
\usepackage{algpseudocode}
\usepackage{physics}
\usepackage{MnSymbol,wasysym}
\usepackage{tikzsymbols}
\usepackage{array,tabularx,multirow}
\usepackage{makecell}
\usepackage{array}

\def\vx{{\bm{x}}}
\def\vy{{\bm{y}}}

\title{Can Structured Data Reduce Epistemic Uncertainty?}

\author{%
Shriram M S$^1$$^*$\and
Sushmitha S$^1$$^*$\and
Gayathri K S$^1$\and
Shahina A$^1$\\
\affiliations
$^1$Department of Information Technology, Sri Sivasubramaniya Nadar College of Engineering\\
\emails
\{shriram2010160, sushmitha2010422, gayathriks, shahinaa\}@ssn.edu.in\\
\textsuperscript{*} Authors with equal contributions.
}

\begin{document}

\maketitle

\begin{abstract}
 In this work, we present a framework that utilizes ontology alignment to improve the learning process of deep learning models. With this approach we show that models fine-tuned using ontologies learn a downstream task at a higher rate with better performance on a sequential classification task compared to the native version of the model. Additionally, we extend our work to showcase how subsumption mappings retrieved during the process of ontology alignment can help enhance Retrieval-Augmented Generation in Large Language Models. The results show that the responses obtained by using subsumption mappings show an increase of 8.97\% in contextual similarity and a 1\% increase in factual accuracy. We also use these scores to define our Hallucination Index and show that this approach reduces hallucination in LLMs by 4.847\%. 
\end{abstract}

\section{Introduction}

In the current era of Large Language Models (LLMs), with an abundance of data, there is always a tricky question to be addressed: Is providing an abundance of data enough to solve complex tasks? The majority of modern-day models are fundamentally probabilistic, which though highly powerful in its way, gives the model only an uncertain output that cannot be reasoned out. This uncertainty is of 2 types, epistemic (EU) and aleatoric (AU), where the former is also called reducible uncertainty, caused due to the lack of knowledge of a model and the latter arises due to randomness in data \cite{HORA1996217}. In this work we aim to show that ontology can be used as a means to reduce epistemic uncertainty in LLMs, wherein we show how additional knowledge acquired through the structural format of an ontology reduces EU. Consider the training data $\{(\vx_i,\vy_i)\}_{i=1}^n \in (\mathcal{X} \times \mathcal{Y})^n$ and our function $F: \mathcal{X} \to \mathcal{Y}$. Our goal is to make the function $F$ learn from such training data that is structured in nature.\par We claim, to tackle these EU's, especially in Language Models where this effect is seen in the form of hallucinations \cite{tonmoy2024comprehensivesurveyhallucinationmitigation}, structured data could be used. In this work, we demonstrate why structured data is more important while attempting to improve the rate at which Language Models learn during the training process. Our experiments also prove how useful structured data is to tune prompts of Generative Language Models. We choose ontologies, which are essentially a collection of named concepts called classes with their properties and relationships. We first attempt to align ontologies by using a pre-trained transformer that is trained to perform sequential classification tasks. The ontologies are subsequently identified as source ontology \(S\) and target ontology \(T\) while aligning, where we can define ontology alignment as a process to identify the semantic correspondence between classes in \(S\) and \(T\). During this process, we consider classes from \(S\) and \(T\) to retrieve equivalence and subsumption mappings, where equivalence implies a pair of same or similar concepts between two classes, one from \(S\) and the other from \(T\), and subsumptions denote the relationship between pair of classes from \(S\) and \(T\), where the class from \(S\) is a superclass of the corresponding class from \(T\) and \textit{vice versa}. \par Our work uses an approach where we make use of subsumptions obtained through ontology mapping, to provide more contextual information to the prompt that is passed to the Language Model. One of the main issues with the current retrieval approaches using Retrieval-Augmented Generation is hallucination, where the model gives out irrelevant, incorrect, and unreal responses. By incorporating subsumptions in the prompt, we ensure hallucination is minimized and the response of the Language Model is more contextually and factually intact. Section \ref{section:4} presents key insights from our experimentation with ontologies in the medical domain, demonstrating how our methodology could be used for quicker training and reducing hallucinations in LLMs. 


\begin{figure*}
    \centering
    \includegraphics[width=1\linewidth, height=4in]{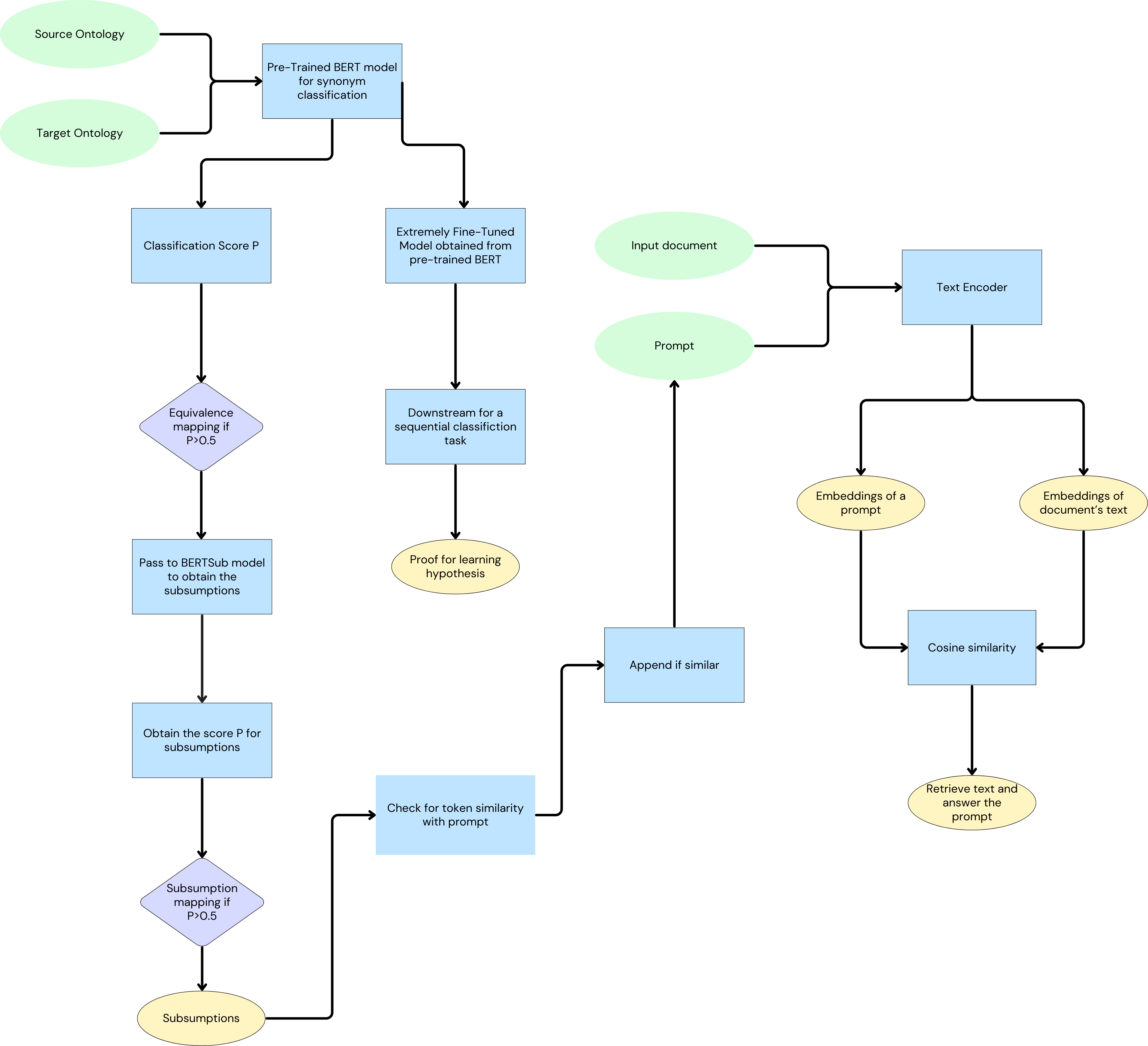}
    \caption{A step-by-step representation of the work}
    \label{fig:1}
\end{figure*}

\section{Literature Survey}

Ontologies and knowledge graph could play a vital role in enhancing the reasoning and contextual capabilities of deep learning models like a transformer, which already learns through an inherent attention mechanism. A lot of research is going on surrounding the topic of ontologies. Yuan He et al. in \cite{he2024deeponto} present a new framework DeepOnto which is a python package developed specifically for ontology engineering. It offers various tools for ontology reasoning, verbalization, normalization, and projection. It also supports ontology alignment as well as the completion tasks using pre-trained LLMs. In \cite{10198434} the authors provide a novel mechanism of ontology alignment using BERTMap, where a fine-tuned BERT was used to predict mappings on a text semantics corpus extracted from ontologies and were then extended and repaired using ontology structure and logic. The model was executed on three biomedical ontology alignment tasks whose results surpassed the existing models. Jiaoyan Chen et al. in \cite{chen2023contextual} introduce BERTSubs, a bert-based model to predict the subsumption relationships in ontologies. In \cite{he2023machine} the authors introduce five new biomedical ontologies extracted from Mondo and UMLS on which ontology alignment tasks were performed and the results were presented. In \cite{he2023exploring} the authors test the use of LLM models such as GPT series and the Flan-T5 for ontology alignment tasks to extract concept equivalence mapping. The conclusion was that LLMs do have the capability to outperform transformer based ontology alignment models such as BERTMap. The authors of \cite{nafar2024teaching} emphasize on the lack of reasoning ability in transformer based models. They introduce a novel approach named probabilistic constraint training where reasoning rules are added as constraints during training thereby providing probabilistic logical reasoning. 
Authors Yu wang et al. in \cite{doi:10.1177/20552076231193213} introduce ‘prompt tuning’, which transforms binary or multiclass classification into mask prediction. This is done using the features that is learned by pre-training the language models. ERNIE- health is used to classify medical texts using prompt tuning and provides promising results. 



The authors of \cite{10113073} introduce an LSTM-based ontology matching to address the biomedical heterogeneity problem. The proposed method enhances alignment quality by incorporating char-embedding for semantic and context information. In \cite{9313345}, the authors propose a concept extraction and classification based on ontology. Bernabé, C.H et al. in \cite{GAYATHRI2020511} discuss a systematic literature mapping to understand how biomedical research can advance with the use of biomedical ontologies. Their results suggest 2 implications, first a method to assess how foundational ontologies are used in biomedical research and second a need for a systematic approach to developing ontologies for better quality and sustainability.
The authors of \cite{app13085060} introduce a hybrid ontological bagging algorithm and an ensemble of ResNet50, VGG16, and Xception models for forest image classification. The approach leverages ontologies to capture semantic relationships, achieving a superior accuracy of 96\% compared to baseline classifiers without ontology. 
Florian Schneider et al. in \cite{10.1145/3587259.3627560} present NLFOA - Natural language focused ontology alignment, that focuses on natural language present in ontology that can be used to process the ontology semantics and the graphical structure. It also showed strong results in zero-shot setup thus reducing the need for human labor to manually align ontologies. The authors of \cite{app10217909} introduce DAEOM novel Ontology Matching framework that combines embedding techniques. It introduces specialized negative sampling for structural relations, takes into account both network structure and ontology terms, and automates the final alignment threshold. Preliminary results on biomedical ontologies show DAEOM's competitiveness with top-ranked systems in terms of F-measure within the Ontology Alignment Evaluation Initiative (OAEI). These works highlight some of the ongoing research in ontology engineering and how they are combined with deep learning approaches.

\section{Methodology}\label{section:3}

We propose a comprehensive framework to help models learn from structured data and hence minimize EU. In this section, we demonstrate how ontologies can be used to reduce EU and help achieve quicker learning. We also explore a novel approach using ontologies to aid RAGs and overcome hallucinations that are often faced by LLMs. Figure \ref{fig:1} shows the overall framework of our work.


\subsection{Using transformers to align ontologies}
In our first step, we perform ontological alignment across two ontologies to obtain equivalence mappings. This is considered a synonyms classification task where we check if class \(c\) in ontology \(O\) and class \(c'\) in ontology \(O'\) are synonymous to each other. We let a pre-trained BERT model, which is essentially a model that is trained to perform sequential classification, predict if a particular class pair  \(\Psi\) is synonymous or not. 

\begin{equation}
   \Psi = (c,c')
\end{equation}
\begin{equation}
   P(c\equiv c') = \{0,1\}
\end{equation}



\newtheorem{hypothesis}{Hypothesis}
\subsection{Usage of the extremely fine-tuned model for sequential classification} \label{section 3.3}

Using the checkpoints obtained through the alignment of ontologies for equivalences mapping, we downstream the task to make the model perform sequential classification. Here we compare both the pre-trained model and the model trained with ontologies, herewith referred to as the extremely fine-tuned model. We train the models over the same dataset to check how well and quickly the models can capture dependencies during training with the same model parameters \(\theta\). Our idea is to define a hypothesis to show how important structured data is for a model. We define our hypothesis as follows: 
\begin{hypothesis} \label{hyp:learning}
Consider two models \(\alpha\) and \(\beta\). Let model \(\alpha\) be an extremely fine-tuned model with additional ability acquired through structured data and model \(\beta\) be the native version of model \(\alpha\).  Both the models are bound to reach an optimal accuracy state \(S\) provided the models run for an arbitrary number of epochs \(\varepsilon\). The factor deciding the better model in such a case would be the rate \(\rho\) at which a model reaches such an \(S\). 
\begin{equation}
    \rho_{S}(\alpha) = X
\end{equation}
\begin{equation}
    \rho_{S}(\beta)=X'
\end{equation}

We say that model \(\alpha\) converges to \(S\) at an epoch \(n\), where \(n < \varepsilon\), giving a rate \(X\) which is quicker than \(X'\), which is the rate at which model \(\beta\) learns at an epoch \(n'\), where \(n' > n\). 
\end{hypothesis}


\subsection{Obtaining subsumption mappings from equivalences}
Using the equivalence mapping obtained, we first attempt to construct a corpus of positive and negative subsumption candidates. We define a positive class pair \(\phi(c1,c_{s}2)\) where \(c1 \equiv c2\) and \(c_{s}2 \sqsubseteq c2\). The BERTSub pipeline (\cite{chen2023contextual}) also creates a negative class pair by replacing \(c_{s}2\) with an arbitary named class pair. We make use of a pre-trained model to predict all the classes \(\phi(c1,c_{s}2)\) in the chosen ontologies and claim \(c_{s}2 \sqsubseteq c1\) when the score \(P(c_{s}2 \sqsubseteq c1) \ge 0.5\). Obtaining such subsumptions through structured data provides us a method to relate concepts across different ontologies, enabling a richer knowledge of the chosen domain for the ontologies.  


\subsection{Prompt tuning using subsumptions to aid RAG}
In this section, we dive into how subsumptions can be incorporated in Language Models ( chatbot, in our case) through prompts to give an enhanced answer.
We train a language model which is provided a document, whose vector embeddings are saved in a vector database. Essentially, this is a RAG module, where the vector database has domain knowledge through the document provided. When a prompt is provided, say \(P\), the model encodes the natural language prompt to give a vector embedding \(\upsilon(P)\). Similarly, the texts of the documents are stored as vectors, giving us \(\upsilon(T_{i})\) where \(T_{i} \in D\). Traditionally, we compute the cosine similarity of these texts 
to retrieve the data for the prompt from the vector database, provided the vector embeddings are similar contextually. The change in our framework comes with what we retrieve from this database, we consider the prompt \(P\) and check if the tokens in \(P\) are the same as \(c1\) where \(c1\) is a class from the source ontology of a subsumption pair  \(\phi(c1,c_{s}2)\). If the token, \(K_{p}\) from the prompt \(P\) is similar to the class \(c1\), we append the class that subsumes \(c1\), that is \(c_{s}2\), to the list of tokens \(K_{p_{i}}\) of \(P\). Then we pass the new prompt \(P'\), which is the detokenized version of  \(K_{p_{i}}\) to the model. This way, the model has new additional concepts to look up, which are contextually related to words in the prompt. This would essentially provide us with an enhanced version of the original prompt with more context being infiltrated into the prompt before it is passed to the model itself.   

\section{Experimentation}\label{section:4}


In this section, we first attempt to prove hypothesis \ref{hyp:learning} by making use of the Symptoms Ontology which contains 1019 classes surrounding symptoms from various systems and functions in the human body as the source ontology, and Clinical Signs and Symptoms Ontology is a collection of 303 classes surrounding various aspects of medical symptoms and signs of diseases as the target ontology. Ontology alignment is performed to obtain equivalence and subsumption mappings. These mappings are a result of the synonymous classification task which is done using a pre-trained version of a BERT-based model. These models are further used to perform sequential classification on the dataset. Using these subsumption mappings we implement the proposed method to infiltrate the prompt to overcome LLM hallucinations and to build on top of the existing RAG frameworks.



To prove hypothesis \ref{hyp:learning} we first used the pre-trained version of BioClinical BERT to perform ontology matching which gives a resultant extremely fine-tuned version of BioClinical BERT that has now been trained on ontologies. This model is now used to perform sequential classification on the MedQuAD-MedicalQnA dataset \cite{Ben_Abacha_2019}. The rate at which the model learns is significantly higher compared to the rate at which a normal BioClinical BERT learns. A similar experimentation was performed using the pre-trained version of bert-base-uncased-yelp-polarity which yielded similar results.

\begin{figure}
    \centering
    \includegraphics[width=0.5\textwidth]{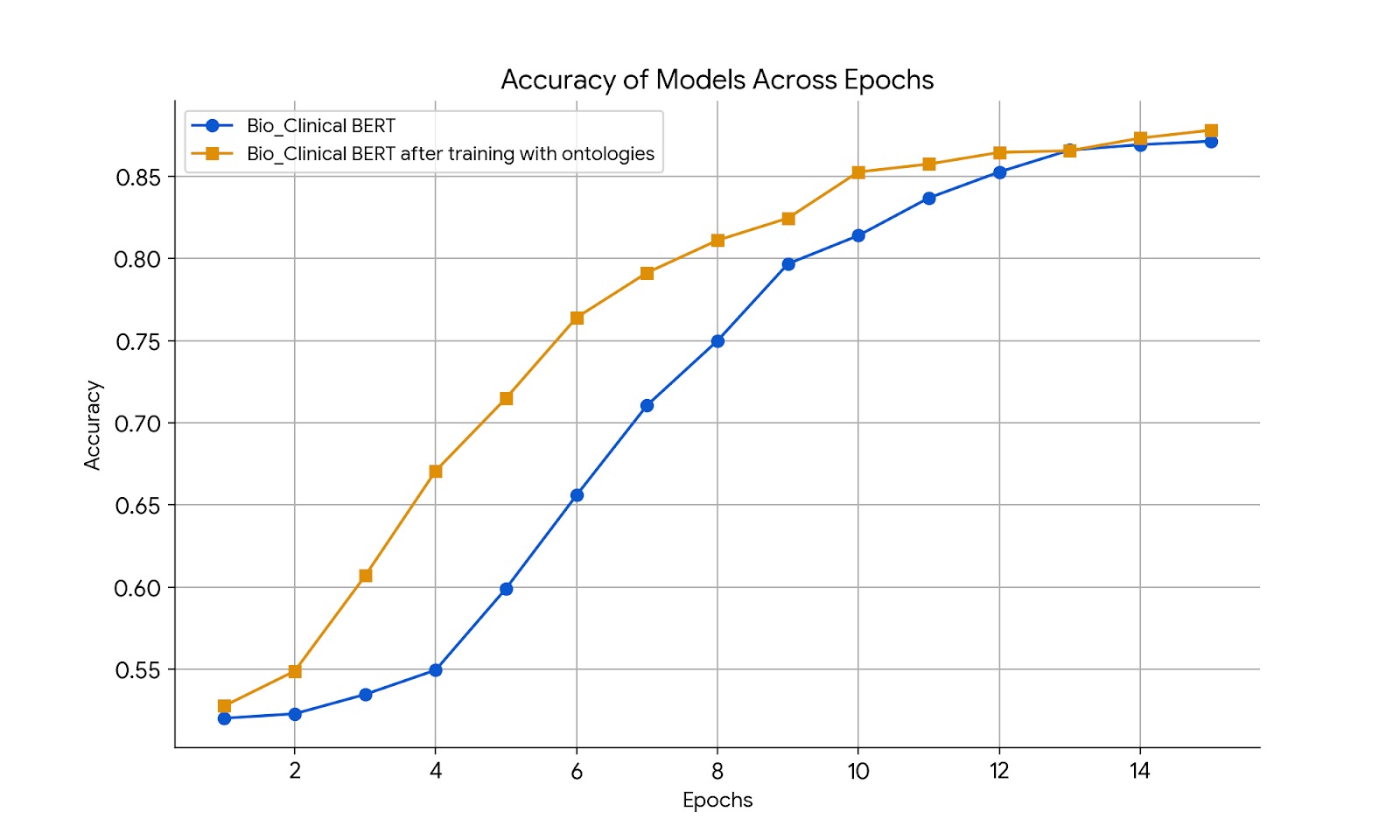} 
    \caption{the accuracies of the extremely fine-tuned and the native models of BioClinical BERT}
    \label{fig:bioacc}
\end{figure}

\begin{figure}
    \centering
    \includegraphics[width=0.5\textwidth]{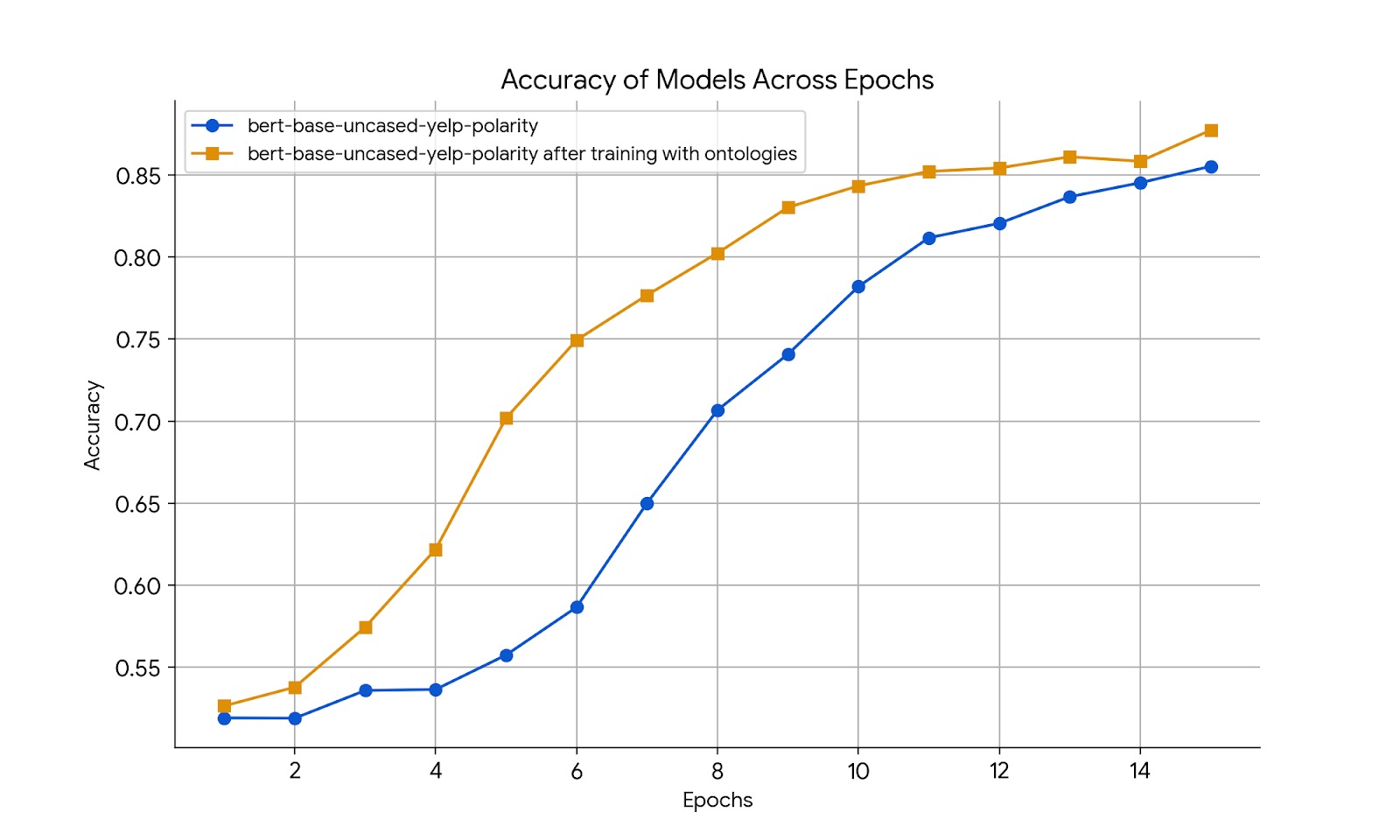} 
    \caption{the accuracies of the extremely fine-tuned and the native models of  bert-base-uncased-yelp-polarity }
    \label{fig:bertacc}
\end{figure}

\begin{figure}
    \centering
    \includegraphics[width=0.5\textwidth]{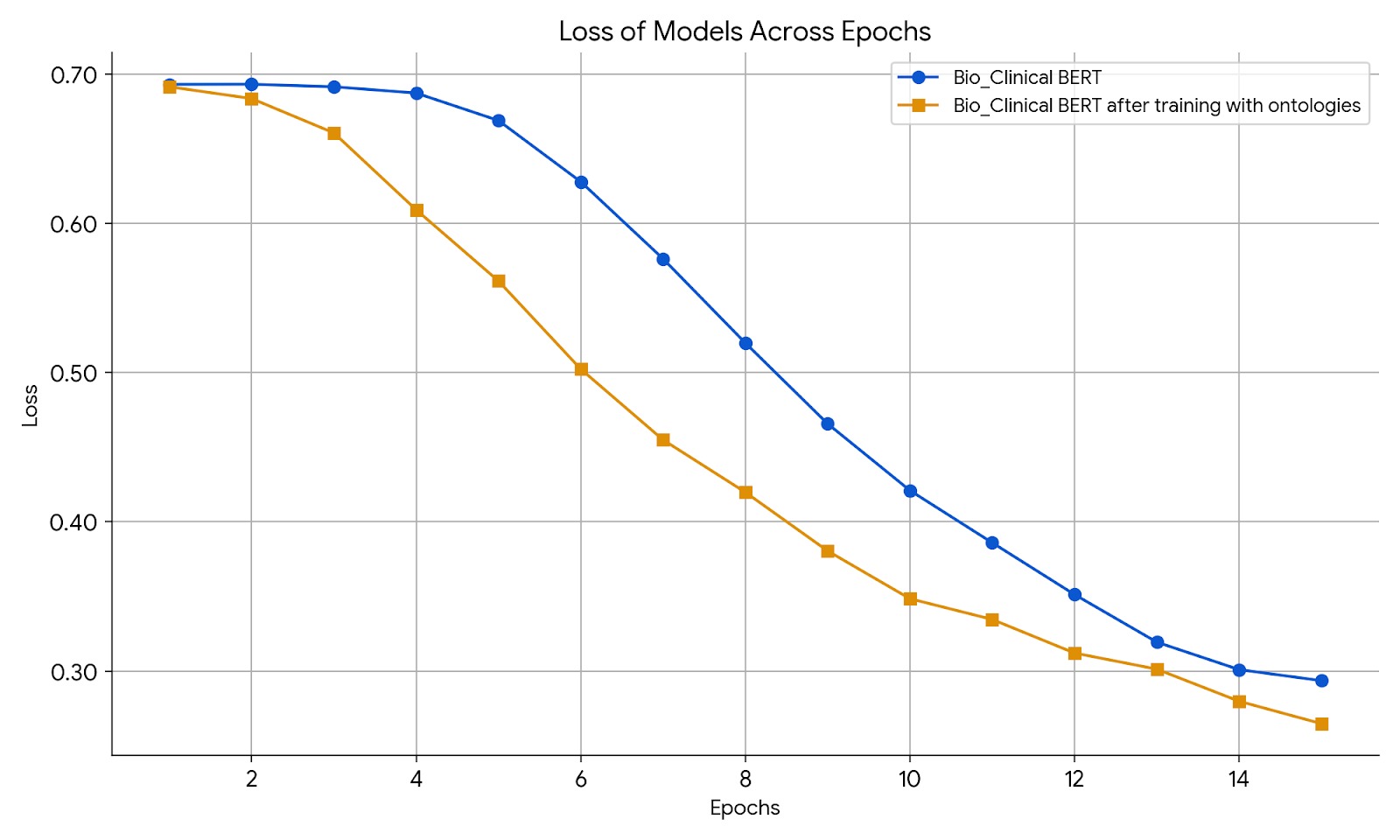} 
    \caption{the loss comparison between the extremely fine-tuned and the native models of BioClinical BERT}
    \label{fig:bioloss}
\end{figure}
\begin{figure}
    \centering
    \includegraphics[width=0.5\textwidth]{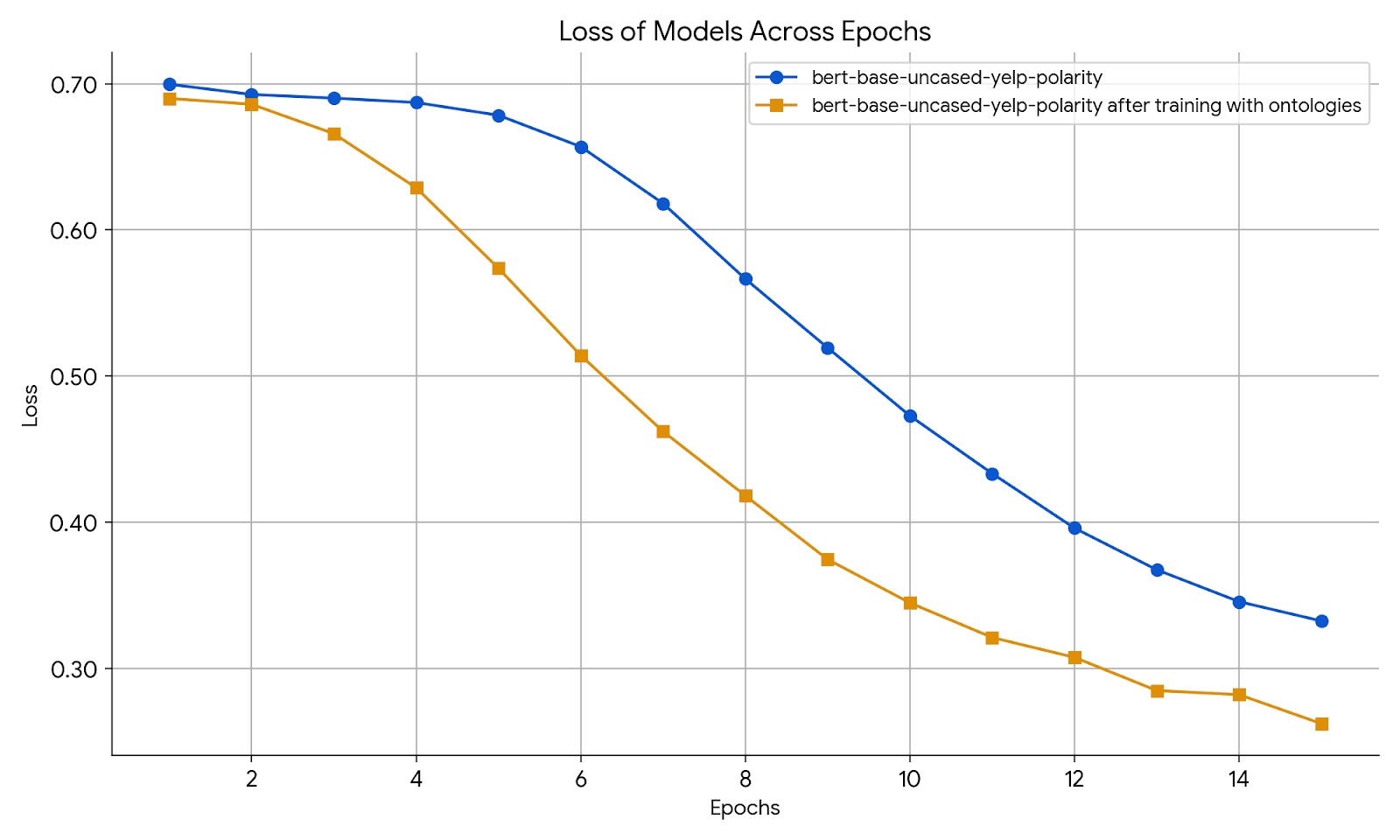} 
    \caption{the loss comparison between the extremely fine-tuned and the native models of  bert-base-uncased-yelp-polarity }
    \label{fig:bertloss}
\end{figure}

The Figure \ref{fig:bioacc} and Figure \ref{fig:bertacc} represent the accuracies of extremely fine-tuned and native models versions of both BioClinical BERT and bert-base-uncased-yelp-polarity models. Analyzing the accuracy after every epoch, it can be inferred that the rate of learning of the model trained through ontology alignment is higher and similarly Figure \ref{fig:bioloss} and Figure \ref{fig:bertloss} show how quickly the model converges to an optimal state. This proves the hypothesis that training a model using structured data enhances the capabilities of the model making it learn at a higher rate.

We create a dictionary that has classes \(c1_{i}\) from the source ontology mapped to its corresponding subsumption mappings \(c_{s_{i}}2\) upon retrieving the class names from the IRIs. As outlined in section 3.5, a chatbot model using the LangChain framework and GPT-3.5-Turbo is created in such a way it utilizes RAG to retrieve the answers for the prompts using cosine similarity. We provide a document that is richly filled with various medical concepts and store the embeddings in a vector database. While the prompts are entered, we parse through tokens in the prompt to check for similar/same tokens as \(c1\) in the prompt. If such a token is found, we append \(c_{s}2\) to the prompt. We pass this new prompt to the model, for it to retrieve the suitable answer.



\begin{table}[H]
\centering
\begin{tabular}{>{\raggedright\arraybackslash}p{0.2\linewidth}>{\raggedleft\arraybackslash}p{0.45\linewidth}>{\raggedleft\arraybackslash}p{0.25\linewidth}}
\toprule
Prompt&  ChatGPT-3.5 turbo with subsumptions& ChatGPT-3.5 turbo\\
\midrule
I have constipation issues what remedies can i take?
&  It is important to consult with a doctor to determine the cause of your constipation. They may recommend dietary changes, increasing the water intake, or using over-the-counter laxatives. They may also suggest lifestyle changes, such as increasing physical activity and managing stress levels. In some cases, prescription medications may be necessary. It is important to address constipation issues promptly to avoid potential complications.& I don't know. Please consult a doctor for proper diagnosis and treatment.\\ 
\bottomrule
\end{tabular}
\caption{Sample prompt and response from the models}
\label{tab:qual}
\end{table}

Table \ref{tab:qual} shows a qualitative comparison of the results obtained from the native GPT-3.5 turbo model and the GPT-3.5-turbo model with subsumption infiltration. 

This way more intricate details are being retrieved from the vector database that stores the knowledge base for the chatbot. Similarly, we provide quantitative analysis to compare the results given out by the models. Our main focus here is to check the extent to which the models are hallucinating, that is, provide a response that is contextually irrelevant, incorrect, and unreal. We define the hallucination index as a score that calculates the extent to which the language model has avoided hallucination. For this, we consider contextual similarity, which is the similarity between the response of the chatbot and the prompt, and factual correctness, which is the similarity between the response of the chatbot and the ground truth present in the document. We formulate the hallucination index by taking an average of contextual similarity and factual correctness.  We consider three different similarity measures, namely cosine similarity, dot product, and Euclidean distance. To do this we first encode the questions, questions with subsumptions, answers from GPT-3.5-Turbo, answers from GPT-3.5-Turbo with subsumptions and the ground truth using a transformer, BERT in our case and use the embedded vectors to calculate the scores. 

 We then compute hallucination index \(H\) as : 
\begin{equation}
    H = \frac{Contextual Similarity + Factual Accuracy}{2}
\end{equation}

Table \ref{tab:consim}  shows how subsumptions help keep intact the context from the prompt, at the same time enhancing it and Table \ref{tab:factacc} shows how closer it is to the lines from the document itself. It is to be noted that cosine similarity and dot product are desired to be a higher value and intuitively, euclidean distance is to be a lesser value. Table \ref{tab:hallind} displays scores that prove how the model tackles hallucination when subsumptions are introduced to the prompt while also minimizing the effect of EU. \textit{Note, in all three tables, cosine similarity is represented as a percentage. } 


\begin{table}
\centering
\begin{tabular}{l>{\raggedleft\arraybackslash}p{0.2\linewidth}>{\raggedleft\arraybackslash}p{0.15\linewidth}}
\toprule
 Contextual Similarity & ChatGPT-3.5 turbo with subsumptions (s) & ChatGPT-3.5 turbo\\
\midrule
Cosine Similarity&  74.8498& 68.4832 \\
Dot Product&  67.376& 59.4732\\
Euclidean Distance&  6.79553& 7.43475 \\
\bottomrule
\end{tabular}
\caption{Contextual similarity scores}
\label{tab:consim}
\end{table}



\begin{table}[H]
\centering
\begin{tabular}{l>{\raggedleft\arraybackslash}p{0.2\linewidth}>{\raggedleft\arraybackslash}p{0.15\linewidth}}
\toprule
Factual Accuracy & ChatGPT-3.5 turbo with subsumptions (s) & ChatGPT-3.5 turbo\\
\midrule
Cosine Similarity&  90.7507& 89.4618\\  
         Dot Product&  71.7345& 71.2838\\ 
         Euclidean Distance&  3.81129& 4.00964\\ 
\bottomrule
\end{tabular}
\caption{Factual Accuracy scores}
\label{tab:factacc}
\end{table}



\begin{table}[H]
\centering
\begin{tabular}{l>{\raggedleft\arraybackslash}p{0.2\linewidth}>{\raggedleft\arraybackslash}p{0.15\linewidth}}
\toprule
Hallucination Index & ChatGPT-3.5 turbo with subsumptions (s) & ChatGPT-3.5 turbo\\
\midrule
Cosine Similarity&  82.8003& 78.9725\\  
         Dot Product&  69.5553& 65.3785\\ 
         Euclidean Distance&  5.30341& 5.72219\\ 
\bottomrule
\end{tabular}
\caption{Hallucination Index}
\label{tab:hallind}
\end{table}

\section{Conclusion and Discussion}
In this work, we prove our hypothesis where we show how a pre-trained model can learn features at a higher rate when fine-tuned on ontologies. Our experimentation assists our hypothesis and opens up a new method to train models. We furthermore display how subsumptions obtained could aid the current RAG framework by enriching the prompt with additional contextual information, hence decreasing the level of hallucinations, thus reducing the effects of epistemic uncertainty. This work could be extended by using a similar framework for multi-modal models, mainly Vision-Language Models, where we can use ontological data to obtain visual information and then utilize this information for further processing.  


\bibliographystyle{kr}
\bibliography{kr-sample}

\end{document}